\documentclass[runningheads]{llncs}

\usepackage{accv}

\usepackage{accvabbrv}

\usepackage{graphicx}
\usepackage{booktabs}
\usepackage{algorithm}
\usepackage{algpseudocode}
\usepackage{nth}

\usepackage[accsupp]{axessibility}  % Improves PDF readability for those with disabilities.

\usepackage{hyperref}

\usepackage{orcidlink}

\begin{document}

\title{REBASE: \underline{Re}ference-\underline{Ba}ckground \underline{S}ubspace \underline{E}limination for Training-Free In-Context Segmentation} 

\titlerunning{Reference Background Subspace Elimination (REBASE)}

\author{Mantha Sai Gopal \orcidlink{0009-0008-4707-9103} \and
Jaison Saji Chacko \orcidlink{0009-0000-2877-4305} \and
Harsh Nandwana \orcidlink{0000-0002-4877-545X} 
\newline
Sandesh Hegde \orcidlink{0009-0009-6983-8614} \and
Debarshi Banerjee \orcidlink{0009-0002-2203-4490} \and
Uma Mahesh \orcidlink{0009-0006-3858-252X}
}

\authorrunning{M.~S.~Gopal et al.}

\institute{CamCom Technologies Private Limited \\ 
\email{\{saigopal.mantha, jaison.saji, harsh.nandwana, \\ sandesh.hegde, debarshi.banerjee, umesh\}@camcom.ai}}

\maketitle

\begin{abstract}
Training-free in-context segmentation enables new object categories to be introduced at inference time from a single annotated reference image, eliminating the retraining and memory overhead of class-incremental learning. Recent approaches achieve this by combining vision foundation models for semantic correspondence with promptable segmentation networks like SAM. However, their performance is fundamentally limited by the quality of the cross-image similarity map; shared contextual backgrounds between the reference and query systematically elevate similarity in non-target regions, degrading prompt localization. We present REBASE, a training-free framework that explicitly suppresses these spurious contextual correspondences. Our method identifies the low-rank background feature subspace from the reference image and project the reference and query features onto its orthogonal complement in closed form, yielding cleaner semantic matching. We then generate positive point prompts using similarity-weighted farthest-point sampling, paired with a refined dense similarity prior. Without any training or parameter updates, our approach establishes a new state of the art among training-free methods on PACO-Part, FSS-1000, and cross-domain datasets such as ISIC2018, demonstrating that explicit background subspace removal is a highly effective principle for one-shot localization. Code is released at: \url{https://github.com/ai-and-lab/rebase}.
\end{abstract}

\begin{figure}[t]
    \centering
    \includegraphics[width=0.9\textwidth]{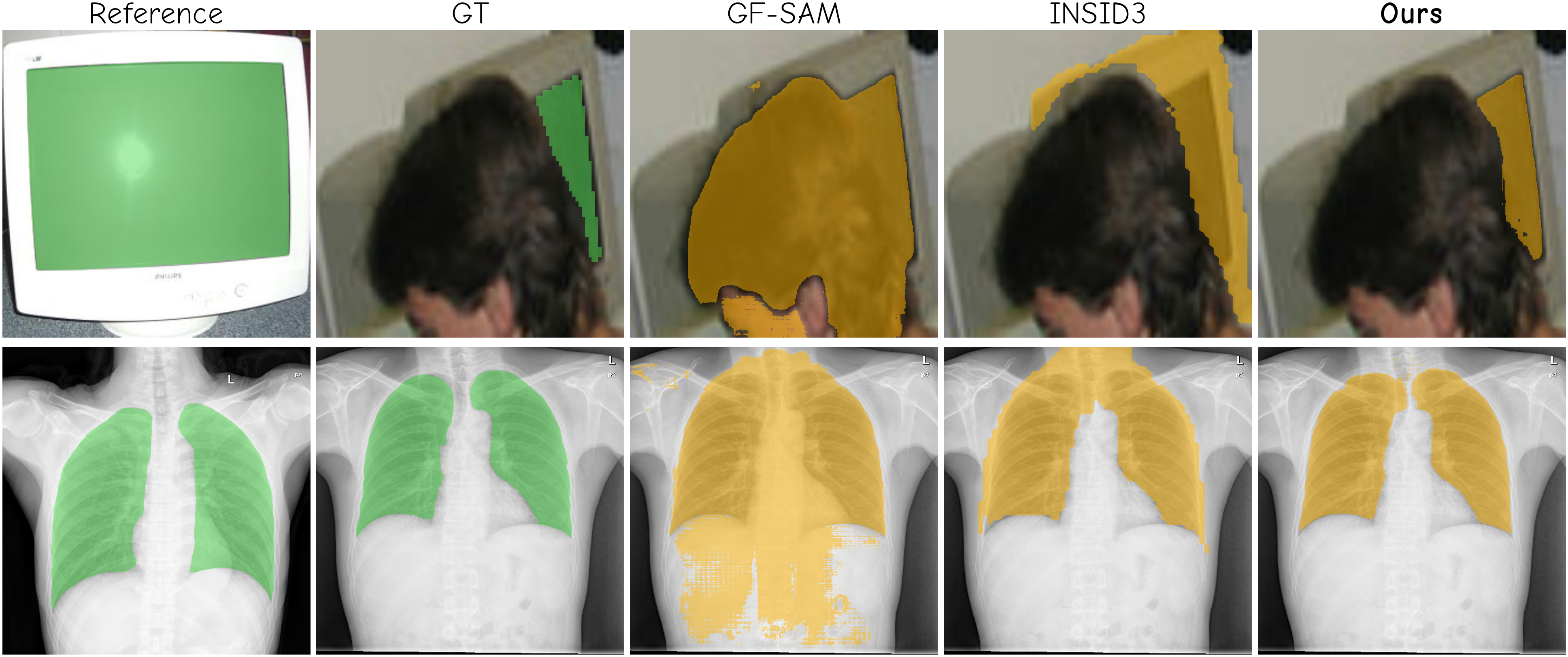}
    \caption{Qualitative examples of our training-free one-shot segmentation method, demonstrating cleaner object boundaries than GF-SAM and INSID3 across two segmentation tasks (top and bottom rows). Each row shows, from left to right, the reference image with its annotation, the query image with its ground truth, and the predictions produced by GF-SAM, INSID3, and \textbf{Ours}.}
    \label{fig:teaser}
\end{figure}

\section{Introduction}
\label{sec:intro}
Real-world deployment of a vision system rarely involves a fixed, closed set of object categories. New instances of interest such as a customer's specific product, a rare medical structure or a domain-specific part arrive continuously after the model has been deployed. \emph{Class-incremental learning} (CIL) is the canonical framework for this setting, wherein a recognition or segmentation model is repeatedly fine-tuned as new classes appear, with mechanisms to mitigate catastrophic forgetting~\cite{li2017lwf,wang2024cilsurvey}. However, in CIL, each new class incurs an optimization pass over potentially the full training set, the storage of either replay data or parameter snapshots, and a non-trivial hyperparameter budget per increment. Moreover, the system is unable to serve a new class until the next fine-tuning step has completed.

Few-shot segmentation on the other hand, offers a structurally different solution. The target class is specified by a small set of annotated reference examples \emph{at inference time}, and the model is tasked with segmenting the target in the query images without requiring any further training or parameter updates~\cite{shaban2017oslsm,min2021hsnet,hong2022vat}. In the one-shot (1-shot) regime, in particular, a \emph{single} reference image with a binary mask must suffice. This setting subsumes the practical desiderata of CIL. New classes can be added on demand, with zero re-training and zero cross-class interference.

The practical realization of this training-free paradigm has been enabled by recent advances in vision foundation models, which naturally decompose the problem into semantic correspondence and geometric localization. Self-supervised vision transformers, such as DINOv2~\cite{oquab2024dinov2}, DINOv3~\cite{simeoni2025dinov3}, learn dense visual representations that exhibit strong semantic consistency across images, thereby facilitating robust zero-shot correspondence estimation. The Segment Anything Model (SAM)~\cite{kirillov2023segment} and its successor, SAM~2~\cite{ravi2024sam2}, establish a general-purpose framework for promptable image segmentation. Trained on over one billion masks, these models accurately convert sparse point or bounding-box prompts into high-quality object masks. However, SAM does not encode an explicit notion of semantic identity. While a prompt on the target object reliably produces an accurate segmentation, the model alone cannot ensure consistent identification of the corresponding object instance across substantial variations in viewpoint, appearance, or scene context. Consequently, the semantic association between the reference and target images must be established externally, typically through dense correspondence models such as DINOv2, with the resulting correspondences serving as prompts for SAM.

PerSAM~\cite{zhang2023personalize} established the canonical training-free recipe for this integration: dense self-supervised features are extracted from the reference and the query, a cosine-similarity map between the query patches and a reference foreground prototype is constructed, and the locations of peak similarity (paired with the complementary minimum as a negative prompt) drive SAM's mask decoder. Subsequent training-free systems, most notably Matcher~\cite{liu2024matcher} and GF-SAM~\cite{cheng2024gfsam} enrich this pipeline with multi-prototype correspondence, controllable mask merging, and explicit positive--negative point alignment. INSID3~\cite{cuttano2026insid3} adopts a similar paradigm but routes the similarity map through hierarchical clustering rather than a SAM-based decoder. Despite the architectural difference, the methods share a common limitation: the discriminative quality of the cross-image similarity map constitutes an upper bound on segmentation accuracy. In natural one-shot data, the reference and query images are independently sampled from the same category but exhibit statistical dependencies in their background distributions. Object categories typically co-occur with characteristic contextual elements. For instance, grass with sheep, sky with birds, indoor scene structure with furniture, or shared anatomical regions across body parts of an animal. The feature directions encoding these contexts are present in both backgrounds, and their contribution to the cosine product with the foreground prototype is positive regardless of whether the corresponding query patch lies on the target. The resulting similarity map is systematically elevated in non-target regions, biasing both the selected point prompts and the dense mask prior away from the true object boundary. The phenomenon is most pronounced for part-level or articulated targets, where the foreground occupies only a small fraction of the patch grid and shared-context patches consequently dominate the top-ranked candidates.

In this paper, we present a training-free pipeline that targets these biases directly. Our contributions are:
\begin{itemize}
\setlength{\itemsep}{2pt}
\item A \textbf{Reference-background subspace elimination} module: an episode-level orthogonal projection $\widetilde{F} = F(I - BB^{\!\top})$ where $B \in \mathbb{R}^{C\times s}$ are the top right singular vectors of the reference's background patch features, applied symmetrically to reference and query DINOv2 features. This isolates and removes scene-context directions shared between the two images.
\item \textbf{Similarity-weighted farthest-point sampling (SW-FPS)} that converts the cleaned similarity map into a spatially dispersed multi-point prompt rather than a single best-similarity point, plus a normalized dense prior injected into SAM's mask-input branch.
\item \textbf{State-of-the-Art Performance and Empirical Validation:} Across five diverse one-shot segmentation benchmarks, our training-free pipeline establishes a new state of the art on ISIC, X-Ray, FSS-1000, and PACO-Part, while achieving competitive performance on PASCAL-Part, all without updating either the feature backbone or the SAM decoder. Figure~\ref{fig:teaser}. shows sample qualitative examples in which our method outperforms previous state-of-the-art methods, by producing more accurate segmentations on challenging examples.
\end{itemize}

\section{Related Work}
\subsection{Foundation Models for Segmentation}
The Segment Anything Model (SAM)~\cite{kirillov2023segment} fundamentally
transformed image segmentation by introducing the first general-purpose
foundation model for the task. Trained on over one billion masks with
class-agnostic supervision, SAM demonstrated that a single promptable model
can generalize across an unprecedented diversity of object categories and
image domains. Its successors, SAM~2 and SAM~3~\cite{ravi2024sam2, sam3},
further extend this paradigm with improved architecture and stronger image
and video segmentation capabilities. Given sparse prompts such as points,
bounding boxes, or coarse masks, these models reliably produce high-quality
object masks while remaining agnostic to semantic category. Parallel efforts
have extended this foundation-model paradigm to specialized domains,
including medical imaging with MedSAM~\cite{ma2024medsam}, high-quality mask
refinement with HQ-SAM~\cite{ke2023hqsam}, and efficient deployment via
distilled variants such as MobileSAM~\cite{zhang2023mobilesam} and
FastSAM~\cite{zhao2023fastsam}.In addition, self-supervised backbones such as DINO~\cite{caron2021dino}, DINOv2~\cite{oquab2024dinov2} and DINOv3~\cite{simeoni2025dinov3} produce rich dense semantic features that serve as strong general-purpose representations for diverse downstream vision tasks.

\subsection{One-Shot and Few-Shot Segmentation}
Few-shot semantic segmentation has a long history of specialized
architectures trained on episodic support--query pairs~\cite{shaban2017oslsm,
min2021hsnet,hong2022vat,zhang2022fptrans}. Early approaches introduced
prototype-based matching~\cite{wang2019panet,liu2020ppnet} and
attention-guided prediction~\cite{zhang2019canet,tian2020pfenet}, later
refined by cycle-consistent transformers~\cite{zhang2021cyctr} and
hypercorrelation squeeze networks~\cite{min2021hsnet}. While these methods
achieve strong performance within their training domains, they require
dataset-specific optimization and lack the flexibility to generalize to
unseen categories without retraining. VRP-SAM~\cite{sun2024vrpsam} learns a visual reference prompt encoder that
translates support annotations into SAM-compatible prompts, and
SINE~\cite{liu2024sine} unifies in-context segmentation across tasks through
a shared prompt interface. Concurrently, visual in-context learning
approaches, such as Painter~\cite{wang2023painter} and
SegGPT~\cite{wang2023seggpt}, formulate segmentation as a unified
image-to-image prediction problem, with follow-ups like
LVM~\cite{bai2024lvm} scaling this paradigm to larger vision sequence
models.

A parallel line of work instead
performs segmentation using frozen, off-the-shelf vision foundation models.
Within this paradigm, PerSAM~\cite{zhang2023personalize} established the
canonical training-free pipeline using DINOv2 for estimating the semantic
correspondence between the reference and query images and generating
positive and negative point prompts used to guide the SAM decoder for target
segmentation. Building on this foundation, Matcher~\cite{liu2024matcher}
incorporates dense bidirectional correspondences and controllable mask
merging, while GF-SAM~\cite{cheng2024gfsam} further introduces explicit
positive--negative point alignment and point--mask clustering.

\subsection{Feature Debiasing for Dense Correspondence}
Mitigating systematic biases in frozen self-supervised representations prior to feature matching has emerged as a highly effective paradigm for training-free dense correspondence. Recent work, such as INSID3~\cite{cuttano2026insid3}, demonstrates that vision transformer features (e.g., DINOv3) harbor strong, disruptive positional priors. Projecting these representations onto the orthogonal complement of a globally estimated, frozen positional basis yields marked improvements in downstream in-context segmentation. Similarly, contemporary methods in the DINOv2 literature identify and suppress position leakage and singular-feature defects via dataset- or corpus-wide calibration~\cite{vandermark2024dvt,sun2024sinder}. Our framework departs from these approaches along two critical axes. First, our orthogonal projection is strictly \emph{reference-conditioned}: the basis for the background subspace is estimated dynamically per episode using the reference image's background patches, rather than relying on static, dataset-wide statistics. Second, the subspace we eliminate encapsulates shared semantic and scene-level context between the support-query pair rather than spatial coordinates. Our approach is therefore conceptually distinct from, and complementary to, existing positional debiasing formulation proposed in INSID3.

\section{Method}
\begin{figure*}[!t]
    \centering
    \includegraphics[width=\textwidth]{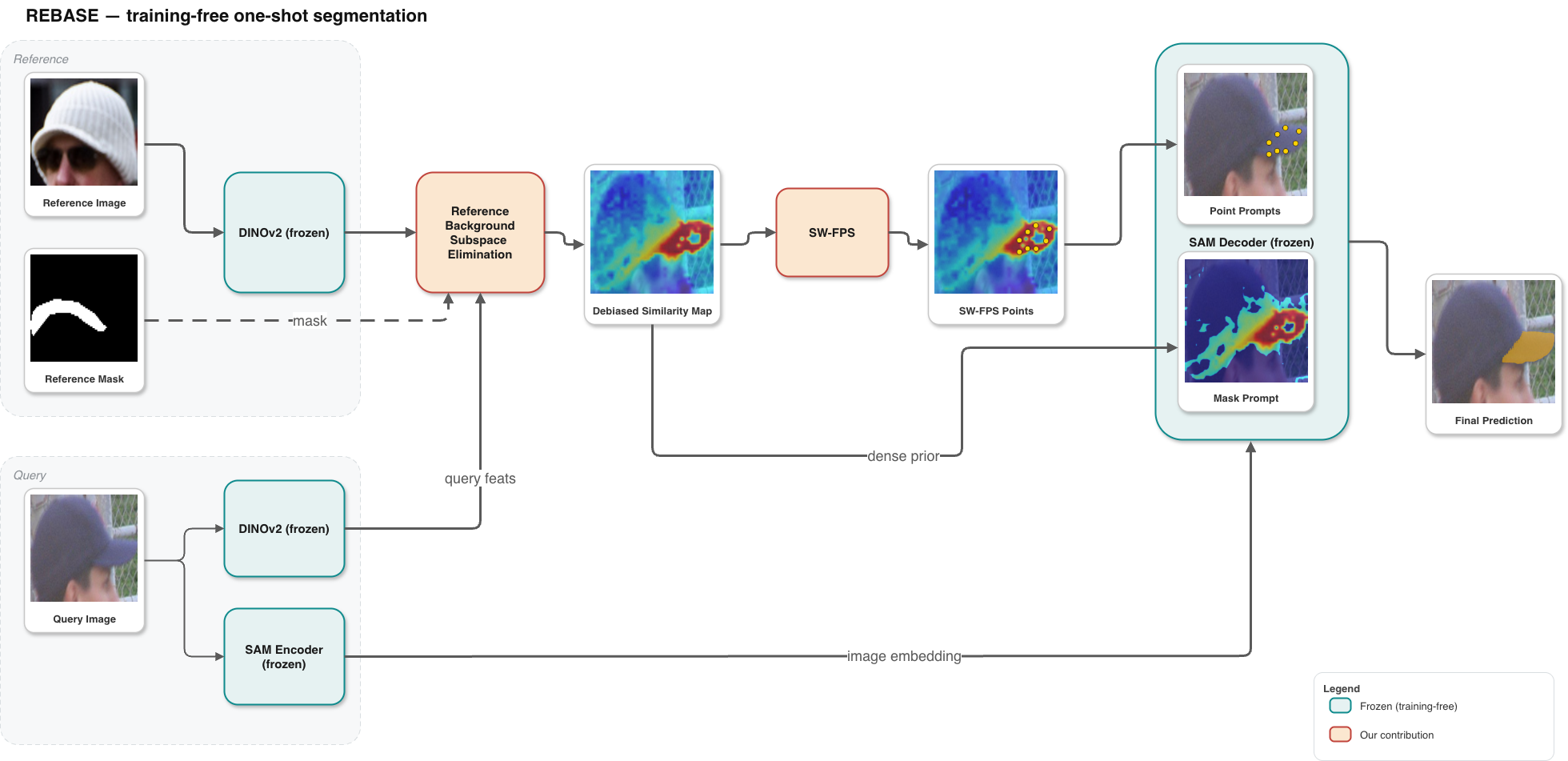}
    \caption{\textbf{Overview of REBASE.}
Given a reference image with its binary mask and a query image, both are encoded by a frozen DINOv2 backbone to extract dense patch features. The reference background subspace is then eliminated, producing a more discriminative cross-image similarity map between the support foreground and query patches. This similarity map is then converted into two complementary conditioning signals for the frozen SAM mask decoder: (i) K spatially diverse point prompts generated via similarity-weighted farthest-point sampling (SW-FPS), and (ii) a dense prior supplied to SAM's mask-input branch. The entire pipeline is training-free, requiring no parameter updates.
}
    \label{fig:method_pipeline}
\end{figure*}

We address one-shot in-context segmentation: given a single reference image $I_R$ with binary mask $M_R$ specifying the target, and a query image $I_Q$, the goal is to predict a binary mask $\widehat{M}_Q$ of the same target in $I_Q$. Our approach, illustrated in Figure~\ref{fig:method_pipeline}. which first applies an episode-level orthogonal projection that removes the dominant background subspace estimated from the reference image from both the reference and query DINOv2 features (Sec.~\ref{sec:method_sbsr}). Using these features, a similarity map is computed (as outlined in Sec.~\ref{sec:method_simmap}). This similarity map is then used to derive spatially dispersed positive prompts using \textit{similarity-weighted farthest-point sampling}, as well as a normalized dense prior for SAM's auxiliary mask-input branch (Secs.~\ref{sec:method_simmap}--\ref{sec:method_dense}). Finally, the generated prompts and dense prior are passed to SAM's frozen mask decoder, which predicts the target segmentation in a single forward pass.

\subsection{Notation and Feature Extraction} \label{sec:method_features}
Let $\Phi : \mathbb{R}^{H\times W\times 3} \rightarrow \mathbb{R}^{h\times w\times C}$ denote a frozen image encoder such as DINOv2~\cite{oquab2024dinov2}. Applying $\Phi$ to the reference and query images yields the feature tensors $F_R, F_Q \in \mathbb{R}^{h\times w\times C}$. To align the binary reference mask $M_R$ with the downsampled patch grid, we construct a soft foreground coverage map $\widetilde{M}_R \in [0,1]^{h\times w}$, where each entry $\widetilde{M}_R(p)$ is the fraction of pixels within the spatial footprint of patch $p$ that belong to the foreground object. In practice, this is computed by area-average resizing $M_R$ from its original resolution to the patch grid $h\times w$, which preserves sub-patch mask detail that a nearest-neighbor assignment would discard.

\subsection{Cross-Image Similarity Map} \label{sec:method_simmap}
Every patch in the reference image with non-zero foreground coverage is treated as an independent prototype, weighted by its coverage following ~\cite{cheng2024gfsam}. Let $\mathcal{F}_R = \{p : \widetilde{M}_R(p) > 0\}$ and let $\widehat{F}_R(p)$, $\widehat{F}_Q(q)$ be $\ell_2$-normalized. The similarity map
\begin{equation} \label{eq:simmap}
    S(q) \;=\; \frac{\sum_{p\in\mathcal{F}_R} \widetilde{M}_R(p)\,\cos\!\bigl(\widehat{F}_R(p),\,\widehat{F}_Q(q)\bigr)}{\sum_{p\in\mathcal{F}_R}\widetilde{M}_R(p)}.
\end{equation}
$S(q)$ is bilinearly upsampled from the patch grid to image resolution before further processing.

\subsection{Similarity-Weighted Farthest-Point Sampling} \label{sec:method_swfps}
\begin{figure}[t]
    \centering
    \includegraphics[width=0.9\textwidth]{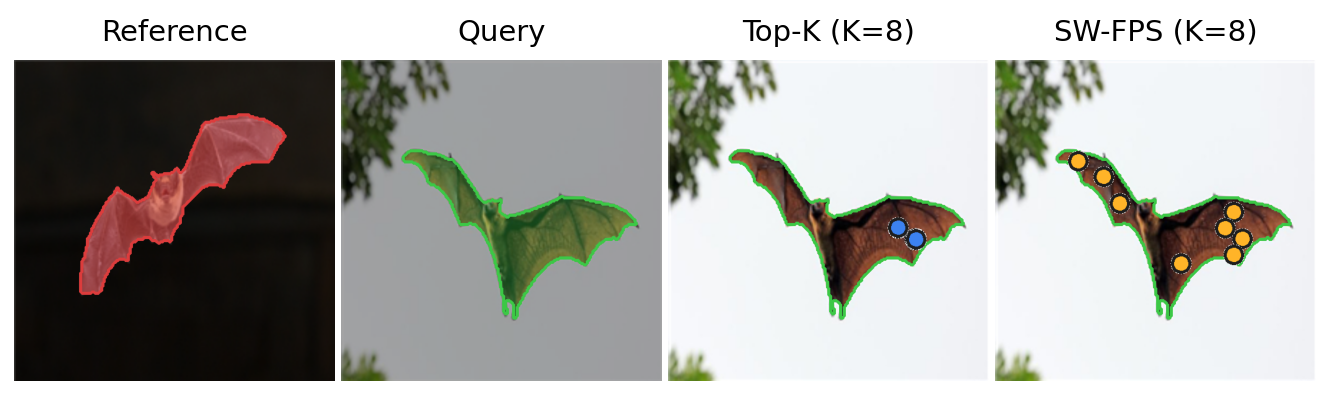}
    \caption{\textbf{Prompt placement: Top-K vs.\ SW-FPS.} Top-K (blue) collapses all $K\!=\!8$ prompts onto the argmax of the similarity map, whereas SW-FPS (orange) disperses the prompts across the target region, guiding SAM toward a more complete target mask. The ground-truth boundary is shown in green.}
    \label{fig:topkvsswfps}
\end{figure}

PerSAM~\cite{zhang2023personalize} derives prompts (a positive prompt and a negative prompt) directly from the similarity map by taking its global maximum (and minimum, as a negative point). While effective when SAM's geometric prior can recover the full object from a single positive prompt, this is insufficient for elongated, articulated, or part-level targets. A naive approach of selecting the top-$K$ points from $S$, collapses prompts onto the same discriminative blob (See Figure \ref{fig:topkvsswfps}).

\begin{algorithm}[t]
\caption{Similarity-Weighted FPS prompt sampler.}
\label{alg:swfps}
\begin{algorithmic}[1]
\Require $S$, $K$, $\alpha$
\State $\mu_+,\sigma_+ \gets \mathrm{mean}(S_{>0}),\,\mathrm{std}(S_{>0})$
\State $\Omega \gets \{p : S(p) \ge \mu_+ + \frac{\sigma_+}{2}\}$
\State $D \gets$ diagonal of the bounding box of $\Omega$
\State $\mathcal{P} \gets \{\arg\max_{p\in\Omega} S(p)\}$
\For{$k = 2,\dots,K$}
    \State $\tilde d(p) \gets \min_{q\in\mathcal{P}}\lVert p-q\rVert_2 / D$
    \State $\tilde s(p) \gets (S(p)-S_{\min})/(S_{\max}-S_{\min})$
    \State $p^\star \gets \arg\max_{p\in\Omega\setminus\mathcal{P}}\,\alpha\tilde d(p) + (1-\alpha)\tilde s(p)$
    \State $\mathcal{P} \gets \mathcal{P}\cup\{p^\star\}$
\EndFor
\State \Return $\mathcal{P}$
\end{algorithmic}
\end{algorithm}

We propose \emph{similarity-weighted farthest-point sampling} (SW-FPS), which selects $K$ point prompts by trading off similarity magnitude against spatial dispersion through a single scalar $\alpha \in [0,1]$. We first restrict the candidate set to patches whose similarity exceeds a confidence threshold,
\begin{equation*}
    \Omega = \{p : S(p) \geq \mu_+ +  \frac{\sigma_+}{2} \},
\end{equation*}
where $\mu_+$ and $\sigma_+$ are the mean and standard deviation of $S$ over its positive support. The first prompt is taken to be the global maximum, $\mathcal{P} = \{\arg\max_{p \in \Omega} S(p)\}$. For each subsequent step $k = 2, \ldots, K$, we select
\begin{equation}
    p^{\star} = \arg\max_{p \in \Omega \setminus \mathcal{P}}
    \; \alpha \, \tilde d(p) + (1-\alpha) \, \tilde s(p),
    \label{eq:swfps}
\end{equation}
where $\tilde s(p) \in [0,1]$ is the min--max normalized similarity over $\Omega$, and $\tilde d(p) \in [0,1]$ is the Euclidean distance from $p$ to its nearest already-selected prompt, normalized by the diagonal of the bounding box of $\Omega$. The two extremes $\alpha = 0$ and $\alpha = 1$ recover top-$K$ selection and pure farthest-point sampling on $\Omega$, respectively. Algorithm~\ref{alg:swfps} states the full procedure.

\subsection{Dense Similarity Prior} \label{sec:method_dense}
Existing training-free personalization methods (PerSAM~\cite{zhang2023personalize}, Matcher~\cite{liu2024matcher}, GF-SAM~\cite{cheng2024gfsam}) drive SAM almost exclusively through sparse point prompts, and dedicate their design effort to selecting a small set of positive (occasionally negative) points from the cross-image similarity map: PerSAM takes the global maximum (and minimum, as a negative), Matcher samples a dispersed set of top-scoring patches, and GF-SAM selects prompts via a graph constructed over high-similarity nodes. SAM's auxiliary mask-input channel, however, accepts a dense low-resolution logit map enabling substantially richer localization cues to the decoder. This channel has so far remained essentially unused by training-free segmentation methods. We argue that this represents a loss of information that the similarity map $S$ encodes.

We therefore inject $S$ directly into SAM's auxiliary mask-input branch as a dense spatial prior. The similarity map is first standardized to zero mean and unit variance to align its dynamic range with that of SAM's mask logits. Its positive support is then interpreted as a coarse foreground estimate, while all remaining locations are assigned a constant negative logit corresponding to the decoder's background regime. The resulting logit map is finally bilinearly resampled to SAM's native $256 \times 256$ mask-input resolution before being passed to the frozen decoder.

\subsection{ Reference-Background Subspace Elimination(REBASE)} \label{sec:method_sbsr}
In self-supervised vision transformers such as DINOv2, patch embeddings are inherently contextualized: global self-attention causes each token to encode both local visual content and information aggregated from the surrounding scene. Consequently, when a reference-query pair exhibits shared contextual attributes such as camera intrinsics, illumination fields, low-frequency surface textures, or co-occurring background semantics, these contextual modes inevitably project onto the reference foreground prototype while remaining active in the query background features.

As a result, irrelevant query regions can exhibit high cosine similarity despite not corresponding to the target object, reducing the reliability of the similarity map. This contextual bias degrades downstream performance along two critical pathways: it compromises the fidelity of the dense prior injected into SAM's auxiliary mask-input branch, and it introduces spatial drift into the positive coordinates localized by the similarity-weighted farthest-point sampling procedure.

We mitigate this with a closed-form, parameter-free, per-episode orthogonal projection of the patch features onto the complement of a low-rank subspace spanned by reference-background patches. From the reference patch grid we construct the index set

\begin{equation}
\mathcal{B}_R = \{p : \widetilde{B}_R(p) \ge \tau_b\}
\label{eq:bgpatches}
\end{equation}
where $\widetilde{B}_R \in [0,1]^{h\times w}$ and  each entry $\widetilde{B}_R(p)$ is the fraction of pixels within the spatial footprint of patch $p$ that belong to the background, stack their DINOv2 features as rows of $X_R \in \mathbb{R}^{|\mathcal{B}_R|\times C}$, and compute its thin SVD,
\begin{equation}
X_R = U_{\,|\mathcal{B}_R|\times r}\, \Sigma_{\,r\times r}\, V_{\,C\times r}^\top, \quad r = \min(|\mathcal{B}_R|, C),
\end{equation}
where the columns of $V \in \mathbb{R}^{C\times r}$ form an orthonormal basis for the row space of $X_R$. The \emph{reference background basis} is the leading block $B = V[:,1:s] \in \mathbb{R}^{C\times s}$, with $B^\top B = I_s$.

Let $P_B = I_C - BB^\top$ denote the orthogonal projector onto $\mathrm{span}(B)^\perp$. We apply $P_B$ symmetrically to every patch feature of both reference and query,
\begin{equation}
\widetilde F_R = F_R P_B, \qquad \widetilde F_Q = F_Q P_B, \label{eq:sbsr}
\end{equation}
so that each patch retains only the component of its feature vector that is orthogonal to the dominant directions of the reference background. Since $B$ is estimated exclusively from the reference background, query patches sharing contextual characteristics with the reference background tend to exhibit large projections onto $\mathrm{span}(B)$ and are consequently attenuated by $P_B$. In contrast, target-related patches typically have weaker projections onto this subspace and are therefore largely preserved.

The background-subspace basis is computed \emph{per episode} from the reference's own background, not once-and-for-all from a noise image or a corpus statistic like in INSID3~\cite{cuttano2026insid3}. The directions $V[:,1{:}s]$ are therefore semantic in nature, encoding the scene context specific to the current reference. Only the rank $s$ is exposed as a hyperparameter.

\section{Experiments}
We evaluated five one-shot segmentation benchmarks that together span three complementary regimes: natural-image semantic segmentation, fine-grained part segmentation, and medical-domain segmentation. For natural-image segmentation, we use FSS-1000~\cite{li2020fss}, which contains 1{,}000 fine-grained object categories evaluated on its standard 240-class test split. For part segmentation, we use PASCAL-Part~\cite{liu2024matcher}, which provides 56 object parts across 15 categories. We also evaluate on PACO-Part~\cite{li2020fss}, containing 303 object parts from 75 categories. For medical-domain segmentation, we use ISIC 2018~\cite{codella2019isic,tschandl2018ham10000} for skin lesion segmentation, and the Chest X-Ray lung dataset~\cite{candemir2013lung,jaeger2014tuberculosis} for lung segmentation.

\paragraph{Implementation Details.}
We use DINOv2 ViT-L/14~\cite{oquab2024dinov2} with frozen weights as the feature extractor and SAM ViT-H~\cite{kirillov2023segment} for mask generation. The input resolutions are set to $518 \times 518$ and $1024 \times 1024$ for DINOv2 and SAM, respectively. The number of positive point prompts is fixed at $K=8$, and the SW-FPS dispersion weight is set to $\alpha=0.5$ across all benchmarks. For the SAM dense-prior branch, we use the z-normalized debiased similarity map and threshold it at the image mean to obtain a candidate foreground region. We then apply a $3 \times 3$ elliptical morphological erosion to suppress thin boundary leakage. Patches outside the eroded region are assigned a background logit of $\ell_{\mathrm{bg}} = -2$. The background subspace basis $B$ is constructed using an adaptive rank, defined as $s = \lceil r \cdot n_{\text{BG}} \rceil$, where $n_{\text{BG}}$ is the number of background patches in the reference image and $r=0.005$. The background patch selection threshold is set to $\tau_b=0.8$ (Eq.~\ref{eq:bgpatches}).

\subsection{Main Results} \label{sec:exp_main}
We compare against two families of prior work. \emph{Fine-tuning methods} learn a segmentation-specific model on in-context data or diffusion features: Painter~\cite{wang2023painter} and SegGPT~\cite{wang2023seggpt} treat segmentation as an image-completion task; SINE~\cite{liu2024simple} and DiffewS~\cite{zhu2024unleashing} condition a diffusion prior on the reference exemplar; and SegIC~\cite{meng2024segic} learns emergent correspondence between the reference and query features. \emph{Training-free methods} freeze a self-supervised encoder and the SAM decoder, and derive point and dense-prior prompts from cross-image similarity. Representative methods include PerSAM~\cite{zhang2023personalize}, Matcher~\cite{liu2024matcher}, GF-SAM~\cite{cheng2024gfsam}. INSID3~\cite{cuttano2026insid3} utilizes only DINOv3-L. Similar to Matcher and GF-SAM, our method belongs to the training-free family and employs the same DINOv2-L backbone and SAM, enabling a fair comparison.

\begin{table*}[t]
    \centering
    \setlength{\tabcolsep}{5pt}
    \renewcommand{\arraystretch}{1.05}
    \small

    \begin{tabular*}{\linewidth}{@{\extracolsep{\fill}}lccccc@{}}
    \toprule
    \textbf{Method} & \textbf{ISIC} & \textbf{X-Ray} & \textbf{FSS-1000} & \textbf{PASCAL-Part} & \textbf{PACO-Part} \\
    \midrule

    \multicolumn{6}{@{}l}{\textit{Fine-tuning}} \\

    Painter \cite{wang2023images}
    & -- & -- & 62.3 & 30.4 & 14.1 \\

    SegGPT \cite{wang2023seggpt}
    & 37.5 & {87.5} & 85.6 & 35.8 & 13.5 \\

    SINE \cite{liu2024simple}
    & 25.8 & 39.8 & -- & 36.2 & 23.3 \\

    DiffewS \cite{zhu2024unleashing}
    & 27.8 & 41.6 & -- & 34.0 & 22.8 \\

    SegIC \cite{meng2024segic}
    & 25.3 & 34.5 & 86.8 & 39.9 & 25.9 \\

    \midrule

    \multicolumn{6}{@{}l}{\textit{Training-free}} \\

    PerSAM \cite{zhang2023personalize}
    & 23.9 & 31.7 & 71.2 & 32.5 & 22.5 \\

    Matcher \cite{liu2024matcher}
    & 38.6 & 70.8 & 87.0 & 42.9 & 34.7 \\

    GF-SAM \cite{cheng2024gfsam}
    & 48.7 & 51.0 & \underline{88.0} & 44.5 & 36.3 \\

    INSID3 \cite{cuttano2026insid3}
    & \underline{54.4} & \underline{78.8} & 83.7$^{*}$ & \textbf{50.5} & \underline{38.7} \\

    \midrule

    \textbf{Ours}
    & \textbf{63.8}
    & \textbf{86.3}
    & \textbf{88.2}
    & \underline{46.6}
    & \textbf{39.3} \\

    \bottomrule
    \end{tabular*}

    \caption{\textbf{Comparison of one-shot segmentation performance across medical (ISIC, Chest X-Ray), generic-object (FSS-1000), and part-level (PASCAL-Part, PACO-Part) benchmarks.} Results are reported as mIoU (\%, $\uparrow$). Best and second-best performances are highlighted in \textbf{bold} and \underline{underline}, respectively.$^{*}$The original INSID3 paper does not report results on FSS-1000; the reported value was obtained by evaluating the authors' publicly released implementation.}
    \label{tab:main_comparison}
\end{table*}

Table~\ref{tab:main_comparison} reports the mIoU results across five benchmarks. Among training-free methods, our approach achieves the best performance on four benchmarks including ISIC, Chest X-Ray, FSS-1000, PACO-Part and ranks second on PASCAL-Part.

\begin{figure}[!t]
\centering
\includegraphics[width=0.7\linewidth]{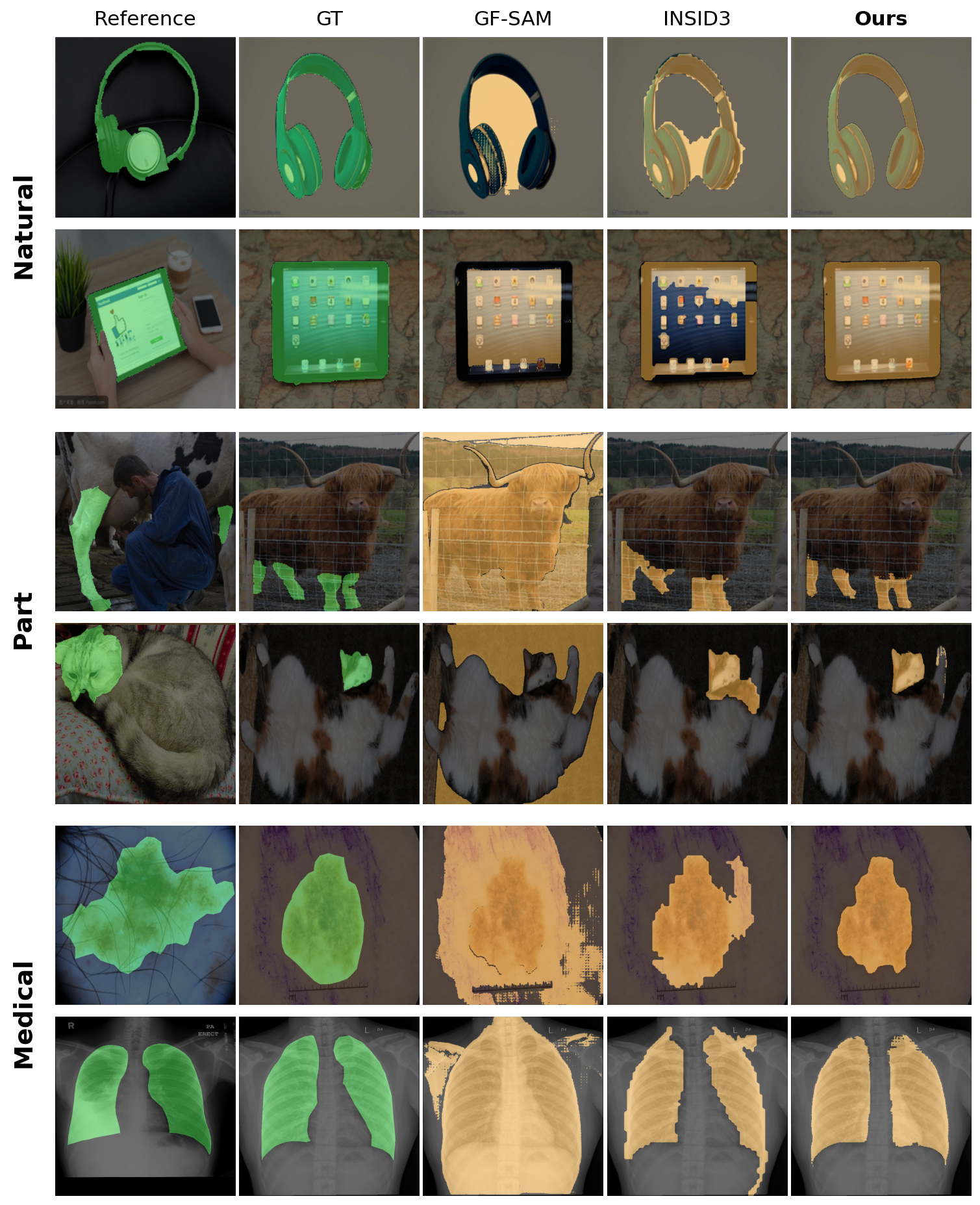}
\caption{\textbf{Qualitative results across the three benchmark categories.} Each row shows, from left to right, the reference image, the query image (both with target overlay), and the predictions of GF-SAM, INSID3, and \textbf{Ours}. Our method produces more accurate segmentation masks with cleaner object boundaries on these representative examples.}
\label{fig:qualitative}
\end{figure}

The largest gains are observed on the medical benchmarks. Our method achieves $63.8\%$ mIoU on ISIC, outperforming the previous best training-free method, INSID3, by $+9.4,\text{pp}$, and reaches $86.3\%$ on Chest X-Ray, improving over INSID3 by $+7.5,\text{pp}$ while remaining $1.2,\text{pp}$ below the fully fine-tuned SegGPT. These improvements align with the motivation of our method: medical images typically exhibit highly structured scene contexts, such as homogeneous skin regions or X-ray backgrounds, making reference-conditioned background-subspace projection particularly effective at suppressing irrelevant context and producing sharper cross-image similarity maps.

On FSS-1000, our method achieves $88.2\%$ mIoU, surpassing GF-SAM ($88.0\%$) by $+0.2,\text{pp}$ and INSID3 ($83.7\%$) by $+4.5,\text{pp}$. Notably, it also outperforms the strongest fine-tuning baselines evaluated on this benchmark, SegGPT and SegIC, without requiring any additional training.

For fine-grained part segmentation, our method attains $39.3\%$ mIoU on PACO-Part, improving over INSID3 ($38.7\%$) by $+0.6,\text{pp}$ and GF-SAM ($36.3\%$) by $+3.0,\text{pp}$. On PASCAL-Part, our method reaches $46.6\%$ mIoU, ranking second behind INSID3 ($50.5\%$). Compared with Matcher and GF-SAM, which also employ the DINOv2-L backbone, our method improves performance by $+3.7,\text{pp}$ and $+2.1,\text{pp}$, respectively. The remaining gap to INSID3 may be partially attributable to its use of the more recent DINOv3-L encoder. Qualitative results across all three benchmark categories are presented in Figure~\ref{fig:qualitative}, illustrating examples where our method produces more accurate segmentations than previous state-of-the-art methods. 

\begin{figure}[!t]
\centering
\includegraphics[width=0.9\linewidth]{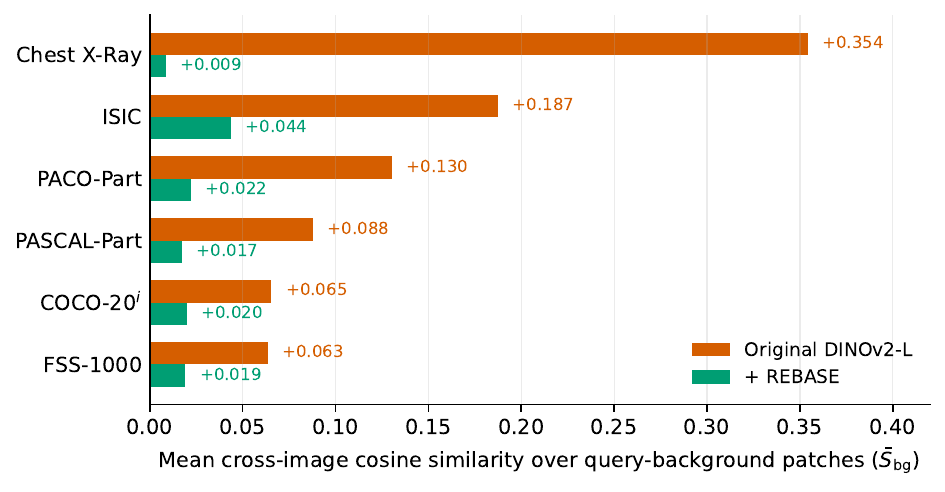}
\caption{Comparison of mean cross-image cosine similarity over query-background patches ($\overline{S}_{bg}$) before and after projection. }
\label{fig:bg_cosine_reduction}
\end{figure}

In addition, we quantitatively demonstrate the elimination of \textit{background bias} by measuring the mean cosine similarity between reference-foreground patches and query-background patches before and after applying REBASE. Figure~\ref{fig:bg_cosine_reduction} shows a consistent and substantial reduction in this similarity across all six benchmarks, with the mean cosine similarity driven to near-zero values after applying REBASE. This indicates that the foreground representations extracted from the reference image no longer exhibit significant similarity to background regions in the query image, providing direct quantitative evidence that REBASE effectively suppresses the spurious cross-image correlations induced by background context.

\subsection{Ablation Study} \label{sec:exp_abl}
We analyze the contribution of each pipeline component on PACO-Part and ISIC datasets. We refer the reader to the Supplementary Material for additional analyses.

\begin{table}[t]
\centering
\setlength{\tabcolsep}{5pt}

\begin{tabular}{p{3.2cm}cc}
\toprule
\textbf{Configuration} & \textbf{PACO-Part} & \textbf{ISIC} \\
\midrule
Vanilla (argmax point) & 29.54 & 38.40 \\
\quad + SW-FPS ($K{=}8$) & 33.23 & 47.47 \\
\quad + Dense Prior& 34.87 & 59.84 \\
\quad + REBASE & 39.28 & 63.77 \\
\bottomrule
\end{tabular}
\caption{\textbf{Ablation of main components of REBASE.} Each row adds one component to the previous. Results are reported as mean mIoU (\%).}
\label{tab:ablation_components}
\end{table}

\paragraph{Main components.}
In Table~\ref{tab:ablation_components}, we investigate the contribution of the different components of REBASE by starting from a vanilla baseline that adopts the argmax point as the prompt. We then replace it with the proposed similarity-weighted farthest-point sampling (SW-FPS), improving performance by $+3.69$,pp on PACO-Part and $+9.07$,pp on ISIC. Next, we add the dense prior, which further improves performance by $+1.64$,pp on PACO-Part and $+12.37$,pp on ISIC. On top of that, we introduce the proposed reference-background-subspace elimiation (REBASE), yielding a further improvement of $+4.41$,pp on PACO-Part and $+3.93$,pp on ISIC, and achieving the best performance of $39.28\%$ and $63.77\%$ mIoU, respectively.

\paragraph{Adaptive background-subspace rank.}
The proposed debiasing step projects the reference and query features onto the orthogonal complement of a rank-$s$ approximation of the background subspace, where the basis $B \in \mathbb{R}^{C \times s}$ is constructed from the top-$s$ right singular vectors of the reference-background feature matrix. Since the number of reference-background patches, $n_{\mathrm{BG}}$, varies substantially across episodes, a fixed rank may exceed the available background evidence in some episodes while capturing only a limited portion of the background subspace in others. We therefore define the rank adaptively as
\begin{equation*}
s = \left\lceil r \cdot n_{\mathrm{BG}} \right\rceil,
\end{equation*}
where $r$ $\in [0,1]$. This allows the rank-$s$ approximation to scale with the amount of available background evidence in each episode, enabling a single value of $r$ to generalize across all benchmarks.

\begin{figure}[t]
\centering
\includegraphics[width=0.8\linewidth]{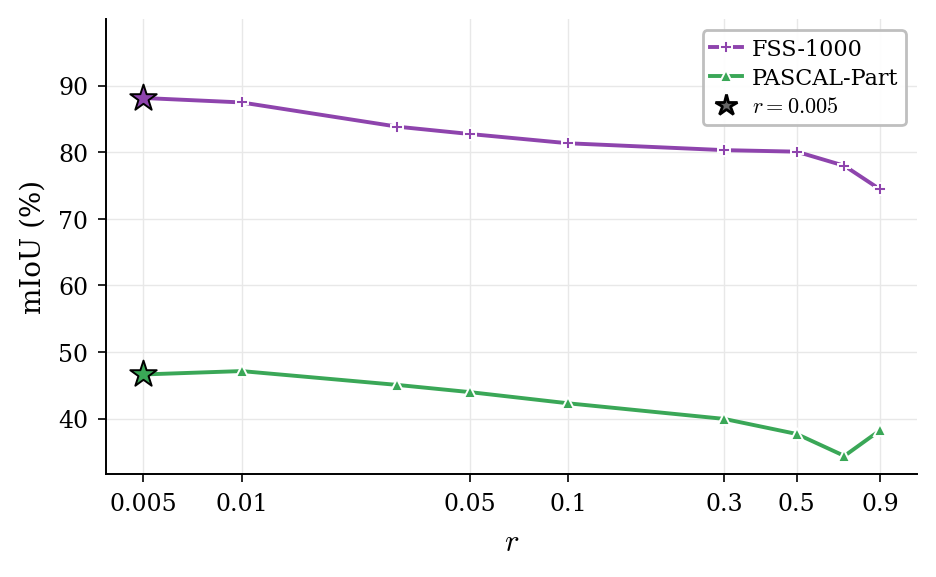}
\caption{\textbf{Sensitivity to the ratio $r$.} mIoU as a function of $r$ (log scale) on FSS-1000 and PASCAL-Part. Performance remains stable in the low-rank regime and progressively degrades as $r$ increases, indicating that only a small number of dominant background singular directions are sufficient for effective debiasing. Our default choice, $r=0.005$ ($\bigstar$), lies at or near the optimum on both benchmarks.}
\label{fig:rank_sweep}
\end{figure}

\paragraph{Empirical low-rank structure of the background subspace.}
The adaptive rank formulation introduces a single hyperparameter $r$, which controls the fraction of reference-background singular directions retained in the subspace basis. Figure~\ref{fig:rank_sweep}. reports mIoU as a function of $r$ over three orders of magnitude on FSS-1000 and PASCAL-Part. Both benchmarks exhibit similar behavior: performance remains stable in the low-rank regime, varying by at most $0.7$\,pp on FSS-1000 and $0.5$\,pp on PASCAL-Part between $r=0.005$ and $r=0.01$, before progressively degrading as additional background singular directions are retained. Increasing $r$ from $0.005$ to $0.9$ results in a total performance drop of $13.7$\,pp on FSS-1000 and $8.4$\,pp on PASCAL-Part. The most pronounced degradation occurs at high ranks ($r>0.5$), suggesting that retaining too many background directions causes the projection to remove target-relevant information in addition to shared scene context. Our default choice of $r=0.005$ (marked $\bigstar$) lies at or near the optimum on both benchmarks.

\section{Conclusion}
\label{sec:conclusion}
In this work, we presented REBASE, a training-free pipeline for in-context segmentation that improves cross-image matching by removing episode-specific background bias from the feature representations. By projecting the reference and query features onto the orthogonal complement of an empirical background subspace, REBASE enhances feature discriminability without requiring either training or test-time optimization. Extensive experiments across natural-image, part-level, and medical benchmarks demonstrate that this simple formulation consistently improves segmentation performance, establishing a new state of the art among training-free methods on four benchmarks while using the same frozen backbone as prior approaches. Furthermore, the proposed adaptive-rank formulation requires only a single global hyperparameter that transfers consistently across all evaluated datasets, making the method both robust and practical. Taken together, these results demonstrate the effectiveness of reference-background subspace elimination as a principled mechanism for improving support-conditioned feature matching. We believe this perspective complements existing foundation-model pipelines and provides a promising direction for developing more robust training-free methods for in-context visual understanding.

\bibliographystyle{splncs04}
\bibliography{main}
\end{document}